\documentclass[conference]{IEEEtran}
\IEEEoverridecommandlockouts
\usepackage[ruled,vlined,linesnumbered]{algorithm2e}
\usepackage{subcaption}
\usepackage{cite}
\usepackage{amsmath,amssymb,amsfonts}
\usepackage{algorithmic}
\usepackage{graphicx}
\usepackage{textcomp}
\usepackage{xcolor}

\makeatletter
\def\ps@IEEEtitlepagestyle{
\def\@oddfoot{\mycopyrightnotice}
\def\@evenfoot{}
}
\def\mycopyrightnotice{
{\footnotesize 978-1-6654-3886-5/21/\$31.00 \copyright 2021 IEEE\hfill} 
\gdef\mycopyrightnotice{}
}

\def\BibTeX{{\rm B\kern-.05em{\sc i\kern-.025em b}\kern-.08em
    T\kern-.1667em\lower.7ex\hbox{E}\kern-.125emX}}

\begin{document}

\title{Configurable Agent With Reward As Input: \\A Play-Style Continuum Generation
}
\author{\IEEEauthorblockN{Pierre Le Pelletier de Woillemont}
\IEEEauthorblockA{\textit{Ubisoft, Player Analytics France} \\
\textit{Sorbonne Université, CNRS, LIP6, F-75005}\\
Paris, France \\
pierre.le-pelletier-de-woillemont@ubisoft.com}
\and
\IEEEauthorblockN{Rémi Labory}
\IEEEauthorblockA{\textit{Ubisoft, Player Analytics France} \\
Montreuil, France \\
remi.labory@ubisoft.com}
\and
\IEEEauthorblockN{Vincent Corruble}
\IEEEauthorblockA{\textit{Sorbonne Université}\\
\textit{CNRS, LIP6, F-75005}\\
Paris, France \\
vincent.corruble@lip6.fr}

}

\maketitle

\pagestyle{plain}

\begin{abstract}
Modern video games are becoming richer and more complex in terms of game mechanics. This complexity allows for the emergence of a wide variety of ways to play the game across the players.
From the point of view of the game designer, this means that one needs to anticipate a lot of different ways the game could be played. 
Machine Learning (ML) could help address this issue. More precisely, Reinforcement Learning is a promising answer to the need of automating video game testing.
In this paper we present a video game environment which lets us define multiple play-styles.
We then introduce \textit{CARI}: a Configurable Agent with Reward as Input. An agent able to simulate a wide continuum range of play-styles. 
It is not constrained to extreme archetypal behaviors like current methods using reward shaping. In addition it achieves this through a single training loop, instead of the usual one loop per play-style.
We compare this novel training approach with the more classic reward shaping approach and conclude that CARI can also outperform the baseline on archetypes generation.
This novel agent could be used to investigate behaviors and balancing during the production of a video game with a realistic amount of training time.
\end{abstract}
\section{Introduction}
\noindent In the board and video game realm, the main goal of Reinforcement Learning (RL) has been to achieve superhuman performances, on board games \cite{AlphaZero}, simple video games \cite{mnih2013playing} or even highly complex ones \cite{AlphaStar}.
However, another interesting application for RL is the pursuit of human decision modeling.
In addition to learning how to win, this approach tries to solve the problem of playing like a human player. Playing like a human is not only an issue of performance level but it adds a new dimension to the problem: the play-style. With modern video games being more complex, the number of different play-styles can be significant. The simulation of various play-style, or play personas \cite{Canossa2009PatternsOP}, has gained interest lately, partly due to their usefulness for the video game designers.

\noindent Having such an agent, that takes into account a play-style, can be of great value to the game designers, as it can be used to test the game in an automated way and give them precious feedback about whether their intentions are actually translated into the game. It can for instance be useful to check if the difficulty of certain parts of the game are as intended. 
Even for game difficulty assessment it is useful to distinguish between performance (i.e. success) and style of play. For example a game can be perceived as very hard for a subset of the players because of their style of play, not because of the game itself. This kind of feedback can be of immense value to the designers.

This is not the first time RL is being used for game balancing : \cite{andrade_hal_01493239} and \cite{andrade_hal_01492622} used RL trained agents to dynamically adjust the difficulty of an opponent. In order to provide the game designer with such an agent for game balancing purposes, one must incorporate the different play-styles in this model. The idea of using trained agent to automate play-testing has been explored in depth by Holmg{\aa}rd et al. through procedural personas generation \cite{holmgard2014Generative}, \cite{holmgard2014Evolving}, \cite{holmgard2018automated}. In their work they mostly focus on archetypal agents and their alignment with players' decisions. In addition they also showcase how those stereotypical play personas can be used to improve content generation.
The novelty of our approach is that through a single training phase, instead of one per play-persona, we produce an agent that is configurable with respect to the reward function in a very granular and interpretable way.
It is done by giving the agent the reward function coefficients as input. We call this approach the Configurable Agent with Reward as Input (\textit{CARI}).
This allows in turn not only to generate caricatural behaviors, but to query any play-style possible and thus create a wide continuum of play-styles to easily be sampled from. It is granular because of the level of details with which a given play-styles can be simulated. It is also interpretable because of the fact that for any play-style, its definition is based on the reward coefficient values, each having very interpretable definition.

In this paper we give first a quick overview of the existing approaches that aim to solve the problem of learning various personas and those aimed at modeling human decision. We then present two approaches to inject information about play-styles into the model without having human data at hand. Both relying on reward shaping, the first relies on training with a fixed reward function, and serves as a baseline. The other approach is based on a dynamic reward function which changes throughout the training. This agent, called CARI, is configurable with regard to the play-style and the performance objectives, not constrained to solely archetypal personas and is obtained through a single training phase.

\section{Background}
\subsection{Play Personas -- Play-Styles}
Canossa and Drachen \cite{Canossa2009PatternsOP} adapt the "persona" framework introduced by Alan Cooper \cite{CooperPersonas} in the field of Human Computer Interaction. They defined personas within a video game as the expression of the persona but within the limited space of a specific video game and called it \textit{play personas}. 
Tychsen and Canossa \cite{PersonasInGame2008} make the distinction between play
mode, play-style and play-personas. It is a distinction based on the level of data aggregation. Play-mode "refers to the behavior of a player with respect to one or a few discrete metrics, within the same overall group or type of metrics". From there play-style is defined as "a set of composite play-mode". And finally play-personas represent the "larger-order patterns that can be defined when a player uses one or more play-styles consistently". Our work here focuses on play-styles.

Not all play-styles perform at the same level.
In the case of a game of infiltration for example, a player who likes to rush in and charge head-on will most likely not perform well.
Those observed play personas result from the analysis of quantitative players' data gathered via telemetrics \cite{PersonasInGame2008}, for instance via clustering algorithms, and indicate the ways the game is played.
Most games today, especially the ones coming from major studios, have built-in tracking mechanisms allowing some form of clustering on the players to be done, based on their behavior within the game. It can be done on key features \cite{StarcraftClustering} or even on sequence-based inputs \cite{DivisionClustering}. This combination of tracking and clustering is done today on many of the big budget games \cite{ClusteringWildlands}. It allows designers to know not only \textit{who} plays their game, but also \textit{how} it is being played.
In this paper, we suppose we do not have any human data available, and focus solely on generating varying and rich play-styles. 
We do so to fit with the assumption that the usefulness of such agent is greater during the production phase than after. And because of this assumption we have to assume that we can not rely on human data.

\subsection{Human Decision Modeling}
To approximate human decision making, Inverse Reinforcement Learning (IRL) \cite{ng2000algorithms} could potentially be used. The 2 main drawbacks of IRL methods in our case, are the absence of available data and the fact that these algorithms assume homogeneity among the trajectories.
This homogeneity assumption goes against the idea of generating varying, and therefore heterogeneous, play-styles.


\noindent There have been many attempts to tackle this idea of play-styles.
To that effect Holmg{\aa}rd et al. use RL \cite{holmgard2014Generative}, evolutionary \cite{holmgard2014Evolving} and MCTS \cite{holmgard2018automated} agents. They demonstrated that reward, fitness or utility function shaping is a great way to achieve variety in the style of play of the agents. Moreover they demonstrated that stereotypical play-styles can be a good low-cost, low-fidelity approach to automated play-test  \cite{holmgard2014Evolving}. They generate 4 play-styles and show that they bear acceptable resemblances with the players on a simple game. They then generate the same 4 play-styles in a more complex environment \cite{holmgard2018automated}, showing the coherence and scalability of their approach.
To better fit with the language used in the video game industry we use the term \textit{archetypes} instead of stereotypical play-styles.
A few issues arise with their approach. The first is the limitation to only 4 archetypes. On more modern and complex games there are more than 4 play-styles and most players do not lie at the edge of the behavioural space but rather are a mix of different aspect of multiple archetypal play-styles, and will likely lie on a continuum between those archetypes. This is simply due to the fact that the more complex the game mechanics, the more numerous the possible \textit{play-modes} and therefore the more numerous the number of play-styles defined from those play-modes.
So, fitting all play-styles into 4 archetypes can become unrealistic as the game mechanics become more complex. Even in \cite{holmgard2014Generative} they note that on one of the play-trace the agent closest to the player is the random one, this could either mean that 4 archetypes is not enough or that relying solely on archetypes is not sufficient.
The second limitations of training separate agents is the reward definition for each of them: how to define the play-style part of the reward function and to efficiently balance it with the performance part. 
Our approach solves both problems by training once and allowing iteration afterward. This allows the practitioner to find the correct balance between play-style and performance to produce meaningful archetypes more easily. It can also be used to generate any behaviors as mix of these archetypes on all dimensions explored in a very granular way.

Holmg{\aa}rd et al. also compared their approach to 'clones' \cite{HolmgClones}, and solve this issue of play-styles variety by fitting one clone per level and per play-style. This solution, while providing good results, is however very impractical in the video game industry where training an agent can take hours or days at a time.
Finally, \cite{kanervisto2020benchmarking} have shown that more classical behavior cloning is very limited in its performance results and that, even though it requires a simulation and a reward function, RL should provide better performance. 

\subsection{Simulation Environment}
There exist already video games that have been used to train RL algorithms, but none that could really be of use for our particular problem of varying play-styles generation. The most famous one is the Atari suite \cite{Bellemare_2013} which, while being very useful due to the vast variety of games, is limited notably by its simplicity: it is not the type of video game where would emerge significantly varied and numerous play-styles. For this to emerge, one needs a more complex game. A good example is StarCraft II \cite{AlphaStar} where the vast complexity of the game offers many different approaches. Unfortunately the main drawback of this game is that it is actually too complex, requiring a vast amount of computation, and very long training time. We needed something complex enough to be able to have different play-styles and some of the challenges that come with training in a complex modern video game, but simple enough to have moderate computation requirements. 
\begin{figure}[htbp]
\centerline{\includegraphics[scale=0.28]{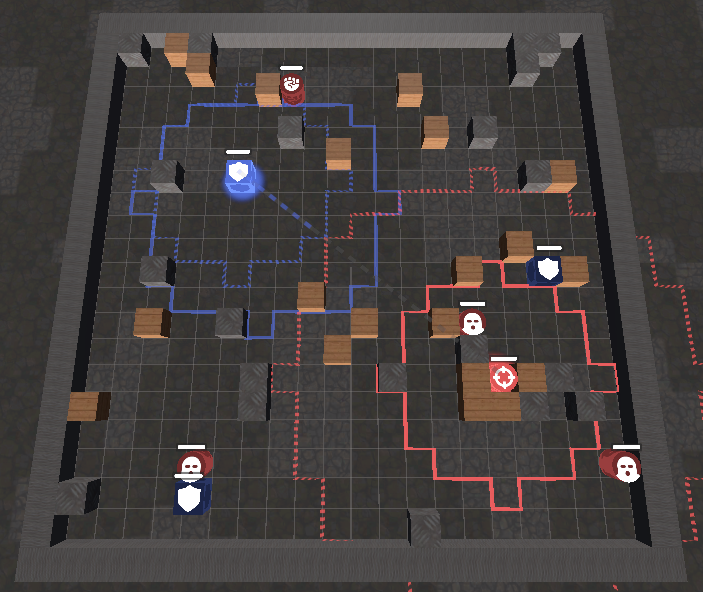}}
\caption{Screenshot of the environment}
\label{fig:Capture_proto}
\end{figure}

\noindent This is the reason why we developed our own video game environment of a discrete, turn based, shooter-strategic video game, see Fig. \ref{fig:Capture_proto}. In this video game 2 teams fight to the death on a 2D square based board with covers laid randomly. One team, composed of 3 heroes, is controlled by the agent trying to learn, and the other team is controlled by hand-crafted decision functions, designed by us like any Non Player Character (NPC) would be in most video games. 

\section{Proposition}
Here we compare two approaches to create agents that can play the game well -- meaning achieve a certain number of wins, but can also play it in a specific fashion. The first, very straightforward one is the simple shaping of the reward so as to orient the agent toward a single given playstyle, it serves as our RL baseline.
CARI, the novel one, on the other hand is based on a reward changing episode to episode.
Allowing for a single model that is configurable with respect to the plays-style in a very granular way.
It relies on having the coefficients of the reward function given as input, in the state description. The two approaches are then tested on their ability to generate 7 archetypal play-styles. The word \textit{archetypes} is used to define these extreme stereotypical behaviors, so that an archetype plays with a caricatural style.
The archetypal play-styles focused on can be each summarized as follows:
\begin{itemize}
    \item Sniper (\textbf{Sn}): prefers the rifle to the knife
    \item Contact (\textbf{C}): prefers the knife to the rifle
    \item Grouped (\textbf{G}): keeps its team tightly grouped together
    \item Scattered (\textbf{Sc}): spread its team as much as possible
    \item Safe (\textbf{Sa}): very adverse to damage taken
    \item DPS (\textbf{D}): focuses on damage inflicted
    \item Win Only (\textbf{W}): cares solely about winning
\end{itemize}

Notice the Win only archetype. While not being a style of play, per say, it is there to serve as a comparative model, to be aware of what is the maximum performance achievable, with a given RL algorithm and given resources at hand.
We focus mainly on 6 archetypes. These 6 archetypes were selected because they are the most commonly observed in such games.

\subsection{Notation}

A RL problem is stated as follows. An agent evolves in an environment. Based on the current state at time $t$: $s_t$, it chooses an action $a_t = \pi(s_t)$. This action $a_t$ is then applied in said environment, yielding a new state $s_{t+1}$ and a reward signal. This reward signal is usually expressed as a function of the state $s$: $R_{\theta}(s) = \sum_{i=1}^{n} r_i * \theta_i(s)$. Where $\theta$ denotes the n-events which are rewarded and $r$ the n-rewards associated with said events. The most common case in zero sum games is to have the agent being rewarded on 2 occasions: +1 in case of a victory and -1 in case of a defeat. Which translates into: $n=2$, $\theta = \{win, lose\}$ and $r = \{+1, -1\}$. 
In this case we go a step further and add $r$ as a parameter of the reward function $R$. The reward function is then expressed as follow: $R_{\theta}(r,s) = \sum_{i=1}^{n} r_i * \theta_i(s)$, with again $r$ the reward coefficients and $\theta$ the events to associate a reward signal with.
The issue of how to define what are these particular events $\{\theta_i\}_{i \in [1:n]}$ rests with the practitioner. 

We introduce the theoretical space $\Phi$ which represents all the possible play-styles that can be defined using the event defined by $\theta$. Meaning that for any play-style $\phi \in \Phi$, $r^{(\phi)}$ are the reward coefficients associated with this play-style.
We also introduce $\Phi_{Arch.} = \{Sn, C, G, Sc, Sa, D, W\} \subset \Phi$ which are the archetypal play-styles defined earlier. So, for example, to simulate a \textit{Sniper} play-style one would select the reward coefficients $r^{(Sn)}$ and use the reward function $R_{\theta}(r^{(Sn)}, .)$ to train a \textit{Sniper} agent. 


The events, rewards and intervals used here are available in Table \ref{tab:reward_coeff_def} and Table \ref{tab:reward_coeff_intervals}. The min and max values for each reward coefficients will serve later on to define the different archetypes. For simplification purpose we consider these intervals as being discrete, each with a step-size of 0.1, and 1 for $r_{Win}$. So that for example: $r_{Win}  \in \{0,1,2, \ldots, 18,19,20\}$ and $r_{Stab} \in \{-1.0, -0.9, \ldots, 2.9, 3.0\}$.

\begin{table}[t]
\begin{center}
\caption{Reward coefficients and events definitions}{}
\centerline{
\begin{tabular}{c|l}
\multicolumn{1}{c|}{\textbf{Notation ($r$)}} & \multicolumn{1}{c}{\textbf{Associated With ($\theta$)}}                                \\ \hline
$r_{Stab}$              & Stabbing an enemy                                                                                             \\
$r_{CvrShooting}$       & Shooting on a cover trying to shoot at an enemy                                                               \\
$r_{HeroShoot}$         & Shooting at an enemy and actually hitting it                                                                  \\
$r_{UsefulShld}$        & An enemy tries to hit a hero but hits its shield instead                                                      \\
$r_{NmyDamage}$         & An enemy actually deals damage to a hero                                                                      \\
$r_{HeroDistance}$      & Average distance between heroes at the end of  every turn                                                     \\
$r_{Win}$               & Wining a game                                                                                                 \\
\end{tabular}
}
\label{tab:reward_coeff_def}
\end{center}
\end{table}
\begin{table}[t]
\begin{center}
\caption{Reward coefficients and their respective ranges}{}
\centerline{
\begin{tabular}{|c|c|c|}
\hline
\textbf{$r$}     & \textbf{Min ($r_{.,min}$)} & \textbf{Max ($r_{.,max}$)} \\ \hline
$r_{Stab}$         & -1.0               & 3.0                \\
$r_{CvrShooting}$ & -2.0               & 1.0                \\
$r_{HeroShot}$       & -1.0               & 2.5                \\
$r_{UsefulShld}$ & -1.0               & 2.5                \\
$r_{NmyDamage}$      & -3.5               & 1.0                \\
$r_{HeroDistance}$ & -3.5               & 3.5                \\
$r_{Win}$           & 0.0                & 20.0               \\ \hline
\end{tabular}
}
\label{tab:reward_coeff_intervals}
\end{center}
\end{table}
It is important to note, at this point, that one needs to select the bounds for each interval based on one's expert knowledge and past experiences. It is a constraint of the approach and should be the focus of future work: develop a methodology to choose these intervals. We do believe this process of interval selection could be automated in the future, using human data. 
It is also important to note that choosing the correct reward coefficients $r^{(\phi)}$ to induce a given play-style $\phi$ is a work that needs to be done by the practitioner at some point, for all RL applications, especially to video games. Taking such large intervals as done here, allows the practitioner to select the correct coefficients after the training is done, and not have to retrain every time they want to change the coefficients. This is amongst the main contributions of this work: train once and iterate afterwards on a already trained and configurable model.

\subsection{Baseline -- One model per play-style}
This approach is the most straightforward. The purpose is to obtain an extremely caricatural behavior, it serves as a RL baseline.
Once the archetypes $\Phi_{Arch.}$ are defined, one needs to map those definitions to corresponding reward coefficients $\{r^{(\phi)}\}_{\phi \in \Phi_{Arch.}}$, or more accurately define for each archetype where each of its reward coefficients should sit in the intervals defined earlier (see Table \ref{tab:reward_coeff_intervals}). Those values, for each of the archetypes, are detailed in Table \ref{tab:archetype_rewards_values}. 

\begin{table}[t]
\begin{center}
\caption{Reward coefficients values for Archetypes play-styles. Note the bold cell, which represent the axes along which a particular archetype is defined.}{}
\centerline{
\begin{tabular}{|c|cc|cc|cc|}
\hline
\textbf{}       & $r^{(Sn)}$ & $r^{(C)}$  & $r^{(Sa)}$ & $r^{(D)}$  & $r^{(G)}$  & $r^{(Sc)}$ \\  \hline
$r_{Stab}$        & \textbf{min}         & \textbf{max}         & $\frac{1}{4}$max   & \textbf{$\frac{3}{4}$max}   & $\frac{1}{2}$max   & $\frac{1}{2}$max   \\[0.75ex] 
$r_{HeroShot}$      & \textbf{max}         & \textbf{min}         & $\frac{1}{4}$max   & \textbf{$\frac{3}{4}$max}  & $\frac{1}{2}$max   & $\frac{1}{2}$max   \\[0.75ex] 
$r_{CvrShooting}$     & \textbf{min}         & max         & max         & max         & max         & max         \\ \hline
$r_{NmyDamage}$      & max         & max         & \textbf{min}         & max         & max         & max         \\ 
$r_{UsefulShld}$    & min         & min         & \textbf{max}         & min         & min         & min         \\ \hline
$r_{HeroDistance}$     & 0           & 0           & 0           & 0           & \textbf{min}         & \textbf{max}         \\ \hline
$r_{Win}$       & $\frac{1}{2}$max & $\frac{1}{2}$max & $\frac{1}{2}$max & $\frac{1}{2}$max & $\frac{1}{2}$max & $\frac{1}{2}$max \\[0.5ex] \hline
\multicolumn{7}{l}{The \textit{min} and \textit{max} values refer to the bounds defined in Table \ref{tab:reward_coeff_intervals}.}
\end{tabular}
}
\label{tab:archetype_rewards_values}
\end{center}
\end{table}

Once the seven (6 play-styles archetypes and 1 win only) $\{r^{(\phi)}\}_{\phi \in \Phi_{Arch.}}$ are defined, one model is trained per $\phi \in \Phi_{Arch.}$. Thus yielding seven agents: six archetypes and one \textit{over-achiever}. We call this approach the \textbf{archetype training} approach.

Note that with this approach one has to carry out one training for each archetype that is targeted. Since \textit{only} 7 archetypes were targeted here, 7 training phases needed to be conducted. It is acceptable in this case, but it is less and less so as the number of archetypes grows, for obvious time and resources constraints. Hence the need to find a way to create similar results but with only one learned model, whatever the number of play-styles.
Moreover, if for some reason the practitioner needs to retrain the model (e.g. due to major changes in the environment) then they need to do so six times, this can become very time and resources consuming. Also if there is, and there usually is, an iteration process to find the correct $r^{(\phi)}$ to produce the desired play-style $\phi$, this process of re-training the agents can become very tedious.

\subsection{CARI -- One Model for all play-styles}
An overview of this approach is available in Algorithm \ref{alg:playstyles_Training}.
Instead of performing multiple trainings, each with fixed reward coefficients $r^{(\phi)}, \phi \in \Phi_{Arch.}$, one single training is carried out with the reward coefficients changing randomly, within their respective intervals (see Table \ref{tab:reward_coeff_intervals}), each episode.
At the beginning of each episode a new value is drawn for each of the reward coefficients (line 5. of Algorithm \ref{alg:playstyles_Training}), and a training episode is run with these rewards coefficients. The 5th line of Algorithm \ref{alg:playstyles_Training} is the equivalent of sampling a random $\phi \in \Phi$ (not just $\Phi_{Arch.})$.
\noindent One needs to add this change to the information available to the agent. To this effect, we simply append these reward coefficients values $r^{(\phi)}$, as is, to the state of the model
Now, the state that the agent takes as input to decide which action to choose is no longer limited to what is going on on the board, but contains also what kind of reward to expect for certain actions, i.e. what the desired behavior is.

\begin{algorithm}[t]
\SetAlgoLined
\KwResult{A single configurable agent (\textit{CARI}) that can adapt its play-style to $r^{(\phi)}, \forall \phi \in \Phi$}
 Define the particular events: $\{\theta_i\}_{i \in [1:n]}$ (see Table \ref{tab:reward_coeff_def})\;
 Define the reward coefficients bounds: $\big\{ [r_{i,min}, r_{i,max}]\big\}_{i \in [1:n]}$ (see Table \ref{tab:reward_coeff_intervals})\;
 Define $L$ the number of episodes to train on\;
 \For{$l=1$ to $L$}{
 ${r}^{(l)} \sim U(\big\{ [r_{i,min}, r_{i,max}]\big\}_{i \in [1:n]})$\;
 Define the following reward function $R_\theta({r}^{(l)},s) = \sum_{i=1}^{n} {r}^{(l)}_i * \theta_i(s)$\;
 Reset environment\;
\While{Episode is not done}{
$s_t = env_t(state)$\;
$s_t = [s_t, {r}^{(l)}]$\;
$a_t = \pi(s_t)$ \;
$s_{t+1}, R_\theta({r}^{(l)},s_{t+1}) = env_t(a_t)$\;
}
}
 \caption{CARI training}
 \label{alg:playstyles_Training}
\end{algorithm}

\noindent By proceeding this way, even though the reward function changes from one episode to the next, it does not actually change with respect to the newly formed state, because the coefficients of this reward function are included in it. By doing so, all the requirements for the algorithm to converge are preserved as the system remains inside a Markov Decision Process (MDP). It is now an MDP in which the objective varies every episode and a model ``aware" of this change. It means that solving this newly defined MDP amounts to solving the original MDP for all play-styles $\phi \in \Phi$.

\noindent This approach learns, through only one training, to behave as any $\phi \in \Phi$, including the 7 archetypes $\Phi_{Arch.}$. It is capable to access a large continuum repertoire of behaviors.
Maybe some players do play like the \textit{Contact} archetype, but it is certain that a lot more players are in between a \textit{Contact} and a \textit{Sniper}, with some degrees of variation. It can also be that some players do play as a combination of a \textit{Contact} and a \textit{Scattered} approach. This method of learning allows to simulate all these cases: the case were the play-style is an archetypal one, the case were the play-style is not an archetypal one and the case in which the play-style is a combination of multiple archetypes.
Thus we name this agent the Configurable Agent with Reward as Input: CARI. It is configurable because the values of the reward coefficient, unlike the rest of the state given to the agent, is given by the practitioner, or even the game designer. By picking a any given $r^{(\phi)}$ one can therefore simulate $\phi$. This is the main contribution of this work: generate a continuum of play-styles $\Phi$ to choose from, the model is no longer constrained by the extreme behavior. The practitioner is also no longer constrained by having to run multiple training to create multiple play-styles, and multiple training for each play-style $\phi$ to find the correct reward function $r^{(\phi)}$.

\section{Experimental Setup}
\subsection{Game Environment}
In this section we describe the game environment in deeper details. As stated before it is a discrete, turn-based, strategic shooter. A team of 3 heroes, controlled by an agent, face off a team of 5 enemies on a 2D square based board, for a maximum of 10 turns. Each hero has the capacity to move, shoot at an enemy, stab an enemy (i.e. melee attack) and apply a shield on itself.
The shield will absorb one attack before disappearing, and then it won’t be available for 2 more turns. Every character has a few basic defining statistics. The most important ones are its health, its range of movement and its range of fire.
So for example during one turn, the agent (controlling the whole hero team) can move around and stab an enemy with hero number 1, then shoot an enemy with hero number 3 and finally put a shield on hero number 2 before skipping the rest of its turn, effectively starting the enemy's turn.

Regarding the enemies, they have available all the actions that the heroes have, apart from the shield which they do not possess. There are 3 types of enemies, distinct based on their \textit{stats}: health, range of move, range of shot and damage. They can be summed up as a high health, low range of movement and high damage (close up fighter), a low health and high range of shot (distance type) and an in-between fighter.
The board is a 20x20 square in which 40 covers are spawn randomly at the beginning of each turn. The spawn position of the heroes and the enemies are also random, allowing for a wide range of possibilities to be encountered during training.

The state returned by the environment is two fold: an image-like segmentation map (called "board") of size 20x20 indicating on each cell what object is inside it (which hero, which enemy, a cover or nothing at all) and an array comprising the rest of the information needed (called "general info"): the number of turns left, the current stats of each hero and each enemy (health, range of movement, range of shot, damage and so on). All in all after some preprocessing operations (in which we will not dive into due to space constraints) this state is of size 3804.
The actions available for the agent are, for each hero, moving in any direction (diagonal included), shooting and stabbing plus the shield action. The agent also has the "end of turn" action, it sums up to an action space of size 61.

\subsection{ACER}
To test our agents, we chose to use the ACER \cite{ACER} algorithm. There are a few reasons why ACER was chosen. First of, it is a discrete action algorithms which suits the problem well. It is an on and off policy algorithm, allowing for both fast convergence and better use of the data generated.
The off-policy part is coupled with a replay buffer, which is prioritized following \cite{schaul2016prioritized}.
Another major reason for choosing ACER is the possibility to run multiple environments in parallel in an asynchronous fashion. This is quite useful for training agents with an environment that is not perfectly stable and could crash.
By "running multiple environments in parallel" we do mean using multiple environments to train the same model, each sharing the same weights and populating the same replay buffer.

The actor-critic aspect of ACER is also an advantage. It will allow, in the future, to be able to use supervised learning for pre-training.
Having those 2 components could also help in the interpretation and explainability of the model \cite{greydanus2018visualizing}.
Say one wishes to use the model to inform game developer about the difficulty of the game. Reporting on the difficulty of the game is one thing, being able to explain \textit{why} is even better. Having an actor-critic architecture allows a wider range of approaches when it comes to explainability.

The neural network architecture used is very straightforward. First, the board is passed through convolutional layers, then it is flatten and concatenated with the general info array and passed through dense layers and finally split in two to output both the policy and Q-value needed. For CARI, the reward function coefficients are added to the state, by simply incorporating them into the general info array as is. The simplicity of this method is also to be noted: it is something that can be applied to any RL problem without much work on the practitioner part.

Note that all rewards are normalized to be bounded by $1$.
For each training phase done, three environments in parallel (each participating in the training of the same model) were used, each running 20,000 training steps, each composed of a transition of 50 time-steps, training the same model. An episode last, on average, around 220 steps, meaning in total, a full model training lasts around $20000 * 50 / 220 \sim 4,500$ episodes. The only reason we run for 20,000 training steps is that it is equivalent to 15 hours given our computational setup, which is a reasonable constraint to aim for in future real-life use-case. We wanted to compare models for which the same number of resources (time and computational power) was allocated. The results shown later are to be read as "given a fixed amount of resources $\ldots$". Our setup is a single computer with a 12 core CPU and a NVIDIA GeForce GTX 1070 GPU.

\section{Evaluation and Results}

\begin{table*}[htbp]
\caption{Mean of the key-metrics for our 3 approaches, averaged over 500 games for each archetype. The cells in bold represent the axis along which the play-style of each archetype is defined.}{}
\begin{center}
\centerline{
\begin{tabular}{|c|ccc|cc||cc|cc||cc|cc||cc|}
\hline
                & \multicolumn{3}{c|}{\textbf{Contact}}         & \multicolumn{2}{c||}{\textbf{Sniper}} & \multicolumn{2}{c|}{\textbf{Grouped}} & \multicolumn{2}{c||}{\textbf{Scattered}} & \multicolumn{2}{c|}{\textbf{Safe}} & \multicolumn{2}{c||}{\textbf{DPS}} & \multicolumn{2}{c|}{\textbf{Win Only}} \\ \hline \hline
\textbf{Agent}  & Heuristic    & B.          & CARI            & B.             & CARI               & B.              & CARI               & B.               &CARI                & B.            & CARI              & B.              & CARI               & B.             & CARI               \\ \hline 
Stabbings       & \textbf{8,9} & \textbf{13,8} & \textbf{13,7} & \textbf{1,3}     & \textbf{0,4}     & 8,5               & 8,1              & 6,1                & 7,1               & 6,4             & 8,5             & \textbf{9,9}      & \textbf{7,4}     & 7,0              & 7,2              \\
Shots           & \textbf{6,3} & \textbf{6,6}  & \textbf{5,8}  & \textbf{12,2}    & \textbf{12,4}    & 8,6               & 10,9             & 8,1                & 9,0               & 11,1            & 10,3            & \textbf{8,6}      & \textbf{10,7}    & 10,7             & 10,2             \\
\%Shots at Covers & 5\%          & 19\%          & 13\%          & \textbf{2\%}     & \textbf{2\%}     & 16\%              & 18\%             & 14\%               & 11\%              & 18\%            & 15\%            & 10\%              & 11\%             & 8\%              & 4\%              \\ \hline
Heroes Dist.    & 5,6          & 7,3           & 5,6           & 9,3              & 8,7              & \textbf{3,0}      & \textbf{4,8}     & \textbf{17,9}      & \textbf{16,6}     & 9,0             & 6,6             & 9,3               & 7,4              & 9,8              & 7,0              \\ \hline
Shields         & 2,6          & 3,3           & 1,8           & 2,5              & 0,8              & 2,5               & 2,0              & 2,8                & 2,1               & \textbf{3,7}    & \textbf{3,1}    & 1,9               & 0,9              & 2,6              & 1,3              \\
Lost HP Heroes  & 81\%         & 91\%          & 92\%          & 74\%             & 86\%             & 72\%              & 84\%             & 81\%               & 82\%              & \textbf{63\%}   & \textbf{50\%}   & 84\%              & 83\%             & 66\%             & 64\%             \\
Lost HP Enemies & 75\%         & 90\%          & 88\%          & 95\%             & 91\%             & 84\%              & 94\%             & 75\%               & 86\%              & 90\%            & 96\%            & \textbf{93\%}     & \textbf{97\%}    & 97\%             & 98\%             \\ \hline \hline
Win             & 21\%         & 53\%          & 45\%          & 77\%             & 67\%             & 44\%              & 72\%             & 26\%               & 41\%              & 61\%            & 88\%            & 75\%              & 82\%             & \textbf{89\%}    & \textbf{92\%}    \\
Lost            & 47\%         & 31\%          & 34\%          & 15\%             & 29\%             & 19\%              & 15\%             & 27\%               & 20\%              & 15\%            & 9\%             & 20\%              & 15\%             & 9\%              & 6\%              \\
Draw            & 32\%         & 17\%          & 21\%          & 9\%              & 3\%              & 38\%              & 14\%             & 48\%               & 39\%              & 25\%            & 3\%             & 6\%               & 2\%              & 3\%              & 2\%              \\ \hline
Turns           & 8,2          & 9,2           & 9,1           & 7,6              & 7,5              & 9,8               & 9,5              & 9,7                & 9,5               & 8,3             & 6,6             & 8,2               & 7,9              & 7,2              & 7,4              \\
Steps           & 173          & 123           & 155           & 112              & 99               & 151               & 98               & 138                & 169               & 215             & 242             & 96                & 97               & 103              & 150             \\ \hline

\multicolumn{15}{l}{\textit{Heuristic} refers to the heuristic developed earlier, \textit{B.} refers to the baseline archetype agents and \textit{CARI} refers to our method.}
\end{tabular}
}
\label{tab:results_all_tracking}
\end{center}
\end{table*}
\subsection{Evaluation Process}
For the 1st approach models, i.e. the baseline archetypes agents, once trained, these models are used to run 500 randomly generated games and track a few key metrics. For the 2nd approach, i.e. the CARI agent, after training, we run the same 500 games and track the same key metrics, with the only difference that for this model we do this tracking once with each of the archetypes reward coefficients $\{r^{(\phi)}\}_{\phi \in \Phi_{Arch.}}$ as input.
This paper aims at comparing the baseline and the CARI agent. It also aims at comparing them with the commonly used way to automate game testing in the video game industry nowadays. To that effect we introduce a simple agent: a \textit{Contact} heuristic. The heuristic agent is here to know what can be achieved in terms of archetype with something that is rule-based, i.e. something that game designers would typically be  doing in the industry to test the game in an automated way.
This heuristic is a simple behavior tree where the heroes try to get as close as possible to the closest enemy, try to stab it and maybe shoot it if it is close enough. It does not take into account the fact that it plays a team of 3 heroes, rather it plays 3 heroes separately. It also does not take into account the previous actions taken, and so there is no real long-term strategy.
Note that the behavior tree controlling the enemies can not be used to control the heroes because the game-play does not allow it: there are more enemies than heroes and only the heroes have access to a shield. We wish to see if using RL can help get a better archetype without all the constraints and limitations of a hard-coded heuristics. 
But the key results of this work is the following: using the already trained CARI we use it to play 20,000 episodes (i.e. 20,000 games). And for each game random reward coefficients were selected.
We display the results as a series of graphs with the horizontal axis being the value of a given reward coefficient and the vertical axis being the key metric associated with said reward coefficient. We will dive deeper into the generation of these graphs later on. These graphs highlight our main contribution: demonstrating our continuously configurable model and allowing the generation and simulation of a very wide variety of play-styles to chose from.

To summarize our evaluation, 3 approaches are compared. The heuristic as a baseline to what is usually used in video games productions. The archetype training agents as a baseline to what is the usual way to generating play-styles through reward shaping. The CARI training which is our contribution and which allows with a single training loop, to be able to behave in many different ways that could be used to investigate behaviors and balancing during the production of a game with a realistic amount of computing time.

There are 2 types of key metrics used: the global ones and the play-style oriented ones. The global ones are: percent of games won, lost or draw (reached the limit of 10 turns), average number of turns, average number of steps. The play-style oriented ones are: average number of stabbings, average number of shots, percent of shots that ended in a cover, the number of shields used, the average distance between heroes, the percent of health lost per the heroes' team and per the enemies' team at the end of the game. The first type of metrics is used to measure pure completion performances, whilst the second type of metrics is used to measure play-style compliance.

\subsection{Results}
All the results of the 3 different approaches introduced earlier are available in Table \ref{tab:results_all_tracking}.
\begin{figure*}[t]
\centerline{
\begin{tabular}{c c c c}
\includegraphics[width=.25\linewidth]{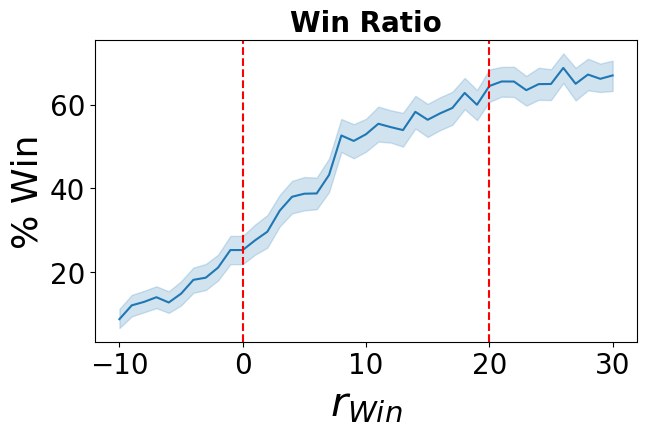}
&
\includegraphics[width=.25\linewidth]{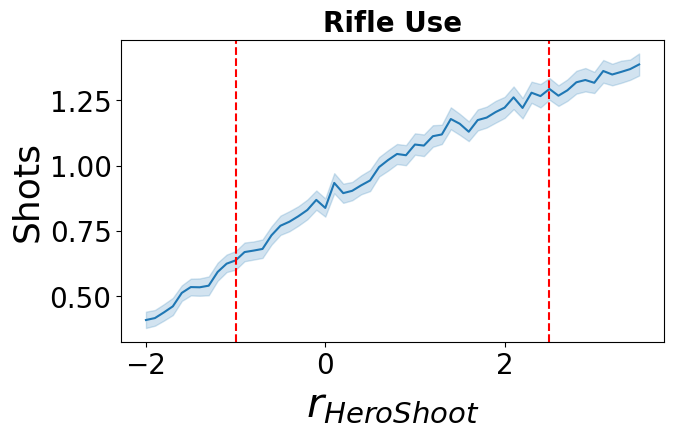}
&
\includegraphics[width=.25\linewidth]{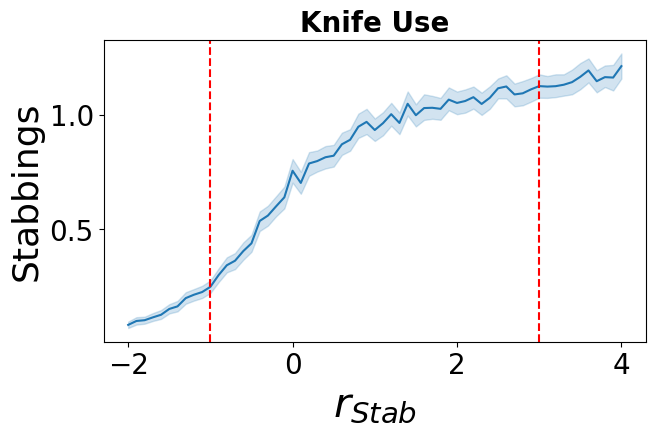}
&
\includegraphics[width=.25\linewidth]{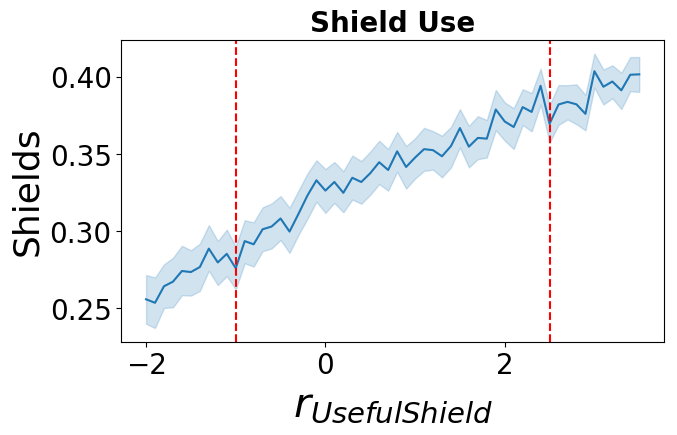}
\\
\multicolumn{4}{l}{ We only display those 4 graphs due to space constraints but we observe similar results on the other reward coefficients}
\end{tabular}
}
\caption{Key-Metrics observed with respect to corresponding reward coefficients using CARI, with their 95\% confidence interval in light blue. The 2 vertical doted red lines represent the bounds of the intervals used during training.}
\label{fig:rdm_rwrd_evolution}
\end{figure*}
\subsubsection{Heuristic} 
The first approach is the most common baseline in video games production teams today: the heuristic one.
The heuristic agent respects the archetype asked from it : it uses the knife more than the rifle. The main drawback is the win ratio ($21\%$), this is due to the fact that if one wishes to hard-code a heuristic it is very challenging to develop one that can deal with multiple objectives -- the two objectives being: playing with a given play-style and also performing well. It is up to the practitioner to decide how much of one objective to sacrifice in favor of the other, thus entering into a tedious iterative loop of trial and error. There is always the risk that, with any big change in the environment, the loop is to be done all over again. 
It ends up being very time consuming, which is what is observed in today's video game productions. 
So the heuristic agent, while respecting key components of the target archetype, does not seem able to actually play the game. It does not display any real strategy in the sense that it plays each hero independently from the others, leading to a somewhat unrealistic way of playing the game. It even loses its advantage of easy implementation when one starts looking at other, more complex play-styles, or if one needs to iterate to find the right balance in the actions to achieve the desired results.

\subsubsection{Baseline Vs CARI}
The first thing to notice is that behaviors generated either by the baseline archetypes or by CARI yield the expected results, both in terms of absolute statistics but also in terms of their respective arrangement with the other behaviors:
\begin{itemize}
    \item The \textit{Sniper} uses the rifle more than anyone else (around 12 Shots per game) while having the lowest number of shots in end in covers (2\%)
    \item The \textit{Contact} uses the knife more than any other (almost 14 stabbings per game)
    \item The \textit{Grouped} and the \textit{Scattered} are respectively the most tightly grouped (around 4 cells of distance between heroes) and most spread out (around 17) teams
    \item The \textit{Safe} is loosing the least amount of HP during games (55\% \textit{Lost HP Heroes} per game)
    \item The \textit{DPS} inflicts the most damages (95\% \textit{Lost HP Enemies} per game)
    \item The \textit{Win Only} wins more than any other (90\% \textit{Wins})
\end{itemize}

\noindent Some of the baseline archetypes have some very low win rates (e.g. 26\% for the \textit{Scattered}). As to why is that, it is quite clear that it is caused mainly by the number of games ending up in draws. This means that the negative reward perceived when getting a timed out (or a loss for that matter) does not outweigh the play style rewards perceived during the game. In other word, the archetype focus on play-style at the expense of the victory. The best examples of that behavior are the Grouped and Scattered archetypes. These 2 perceive a reward at the end of each turn based on the average distance between all heroes, and it seems that they would rather finish on a draw and gather those end-turn reward rather than win and renounce gathering those additional end-turn reward. Waiting for the last turn to win the game is not something they were able to learn consistently.
So, having better win ratio would require tweaking the reward function and thus entering in the tedious iterative loop mentioned earlier. The other drawback of this method is that 7 different models had to be trained, which translates into $7*15=105$ hours of training.

It seems that CARI solves this issue. Let us take the example of the \textit{Grouped}. It goes from $44\%$ win rate to $72\%$. But the average number of turns is only reduced a little: from $9.8$ to $9.5$. That means that the agent is taking advantage of both the end-turn reward and the end-game reward. It is able to learn such things because, and this is intrinsic to this approach: all possible play-styles are trained simultaneously. Meaning that the agent is well aware of what the end-turn and end-game rewards are, simply because it was encouraged to perceive one and/or the other depending on the episode.
There still remains some imperfections; if we look at its counterpart: the \textit{Scattered}. It goes from $26\%$ to $41\%$ win rates. While still being an improvement it remains quite low. This could be explained by the fact that such a high \textit{Average Heroes Dist.} of around $17$ is just intrinsically detrimental to the winning abilities of the agent. Spreading one's team that much does not allow for any meaningful winning strategies. Lowering the $r_{HeroDistance}$ or increasing the $r_{Win}$ might solve this issue. CARI allows for such tweak without needing any retraining of any kind, making this iteration loop to find the proper $r^{(Sc)}$ much faster.

\subsubsection{CARI's Behavioral Continuum}

The other main motivation, beyond archetypal behaviors, is to be able to access the play-styles in-between those archetypes. While extreme reward coefficients $\{r^{(\phi)}\}_{\phi \in \Phi_{Arch.}}$ were used to produce Table \ref{tab:results_all_tracking}, smaller rewards coefficients could be used and so create varying degrees of these play-styles. This is usually the case with human players: there are a few extreme archetypes and a lot more people in between. With this sort of model we can get both of those groups.
To that effect we report additional results. To obtain those additional results, the already trained CARI agent was used to play 20,000 different randomly generated games. Each of these levels were played using a different randomly generated reward function, by sampling as in line 5 of Algorithm \ref{alg:playstyles_Training}, and keeping it constant during the level.
The reason random reward functions are drawn instead of looping over all possible combinations is because, our reward coefficients intervals make up for a total of $4.32*10^{15}$ possible combinations.
It is important to note that there is no training of any kind going on here. The results generated through this process are showed in Figure \ref{fig:rdm_rwrd_evolution}. The metrics represented here are normalized by the number of turns. For example, it is not the number of shots in a game, but in fact the average number of shots per turn.

It is quite clear that CARI has learned to adapt its behavior, its style of play, depending on which reward coefficient it is subject to. The most interesting one is the win ratio. Even on something very abstract such as winning or loosing, it has learned how to adapt to what is asked of it. The other thing to note here is the 2 red doted lines on each graph representing the intervals seen during training by the model. As we can see the model learns to generalize quite well on any dimension to previously unseen reward coefficients values and to adapt its behavior accordingly. These results are very encouraging as they show that one can train a model to learn many different ways to play the game at once. All this in just one training.
It also means that one could possibly increase the number of play-styles one wishes to obtain without increasing the amount of training time needed.
This is a tremendous gain both in terms of time and computation power. It also requires no human data, making it feasible to apply to any video game in production, even during the early stages.

It does however raise a few challenges. The two main ones are: how to choose the interval of the reward coefficients and how to then choose in these intervals which values to select as to generate a human-like agent. Many different avenues might be explored. For example it could be that training a model in a supervised fashion to predict which reward coefficients are used based on the metrics observed might yield interesting results, and allow no human data in the training loop.
Another avenue worth exploring are Quality-Diversity evolution algorithms \cite{qualityDiversity}, which could be used with a CARI agent to generate high-performing archetypes easily.
This model, allowing to query behaviour from an already trained model, can help speed up many algorithm of the sort by moving the training of the RL algorithm outside the main iteration loop, for example it could help achieves the same results as in \cite{ijcai2020-466} without having to train a deep RL algorithm at every iterations.

\section{Conclusion}
We have shown the benefits of using RL over heuristics to create stereotypical agents, and of using varying reward functions across episodes rather than across training phases to create a continuum of play-style through one single training phase. This configurable agent allows better sampling in the space of possible play-styles and does not require any human data, which makes this approach realistic with the constraints of the video game industry. Moreover, our approach is agnostic to the RL algorithm and to the environment used meaning it can easily be applied on many policy optimization problems.

While effective, the approach presented here still requires expert knowledge about the environment, as many RL applications. Using players data could alleviate this problem in the future. Using players data could also help understand whether players' play-styles are more of a continuum, between different archetypes, or more distinct clusters. With the CARI training approach, whether it is the former or the latter, the model is equipped to simulate both since it can simulate the continuum and it can easily be constrained to simulate given clusters.


\bibliographystyle{IEEEtran}
\bibliography{IEEEabrv, conference_101719}

\end{document}